\pdfoutput=1
\documentclass[]{article}
\usepackage{proceed2e}

\usepackage{times}
\usepackage{graphicx} 
\usepackage{subfigure} 

\usepackage{url}
\usepackage[numbers]{natbib}

\usepackage{algorithm}
\usepackage{algorithmic}

\usepackage{amsmath, amsthm, amssymb}
\usepackage{multirow}

\newenvironment{definition}[1][Definition]{\begin{trivlist}
\item[\hskip \labelsep {\bfseries #1}]}{\end{trivlist}}

\title{The Wreath Process: A totally generative model of geometric shape based on nested symmetries}

\author{} 

\author{ {\bf Diana L. Borsa\thanks{Part of this work was undertaken while DB was an intern at Microsoft Research Cambridge}} \\
Computer Science, CSML \\
University College London\\
\And
{\bf Thore Graepel \thanks{TG was at Microsoft Research Cambridge during this work}}  \\
 Google Deepmind       \\
 University College London \\
\And
{\bf Andrew Gordon}   \\
 Microsoft Research \\
 University of Edinburgh \\
}

\begin{document}
\newcommand{\W}{\mathop{\textcircled{w}}}
\newcommand{\tshape}{{\cal S}}
\newcommand{\troot}{{\tt root}}
\newcommand{\tgroup}{{\tt group}}
\newcommand{\toccup}{{\tt occ}}
\newcommand{\ttrafo}{{\tt trafo}}
\newcommand{\taxis}{{\tt axis}}
\newcommand{\ttranslate}{{\tt Trans}}
\newcommand{\trotate}{{\tt Rot}}
\newcommand{\tmirror}{{\tt Mirror}}
\newcommand{\tstretch}{{\tt Stretch}}
\newcommand{\tscale}{{\tt Scale}}
\newcommand{\tline}{{\tt Line}}
\newcommand{\tcircle}{{\tt Circle}}
\newcommand{\torigin}{{\tt Origin}}
\newcommand{\tx}{{\tt X}}
\newcommand{\ty}{{\tt Y}}

\newcommand{\displaycomment}[1]{\marginpar{\raggedright\scriptsize{#1}}}
\newcommand{\todonote}[2]{\displaycomment{{\bf#1}:{#2}}\typeout{TODO #1: (\thepage) #2}}
\newcommand{\ADG}[1]{\todonote{Andy}{#1}}
\newcommand{\DB}[1]{\todonote{Diana}{#1}}

\maketitle

\begin{abstract}

We consider the problem of modelling noisy but highly symmetric shapes that can be viewed as hierarchies of whole-part relationships in which higher level objects are composed of transformed collections of lower level objects. To this end, we propose the stochastic wreath process, a fully generative probabilistic model of drawings. Following Leyton's "Generative Theory of Shape", we represent shapes as sequences of transformation groups composed through a wreath product.

This representation emphasizes the maximization of transfer --- the idea that the most compact and meaningful representation of a given shape is achieved by maximizing the re-use of existing building blocks or parts.

The proposed stochastic wreath process extends Leyton's theory by defining a probability distribution over geometric shapes in terms of noise processes that are aligned with the generative group structure of the shape. We propose an inference scheme for recovering the generative history of given images in terms of the wreath process using reversible jump Markov chain Monte Carlo methods and Approximate Bayesian Computation. In the context of sketching we demonstrate the feasibility and limitations of this approach on model-generated and real data.  

\end{abstract}

\section{INTRODUCTION}
\label{intro}
A substantial part of human knowledge and reasoning is based on the idea of a part-whole hierarchy in which an entity at a given level represents a part of a whole at a higher level and is itself composed of parts defined at a lower level. This notion is crucial to the natural sciences (elementary particle, atom, molecule, cell, organ, organism, etc), but is also at work in the structure of documents (letter, word, sentence, paragraph, section, chapter, book), music (single note, chord, chord progression, piece etc.), or architecture (wall, room, wing, building). Even plans of action are often best represented in terms of hierarchies of goals and subgoals. While it may be debatable if reality itself can be said to exhibit this kind of structure, it is clear that the human mind frequently resorts to the principle of hierarchy in organizing complex structure. In this paper, we take a mathematical formulation of this idea provided by Michael Leyton \cite{Leyton01Generative} based on the group-theoretic notion of wreath product and show how one can build probabilistic models of shape that discover a hierarchical generative representation - providing the basis for understanding and manipulating the shape.
  
The view of computer vision as inverse computer graphics is very elegant and has a long history in the field. The key idea is to define a graphics language to describe the generative process for creating a class of images and---given an image---to infer its generative history in terms of that language. To account for irregularities, noise, and ambiguity, we define a stochastic rendering process that mitigates the rift between the platonic graphics language and the reality of the image \cite{Mansinghka2013}.While it is difficult to make this paradigm work for general classes of images, we focus on the special case of hand-drawn, highly symmetric geometric sketches. 

What would be a good graphics language to describe geometric sketches? Typical drawing tools (including those in PowerPoint) provide graphics primitives such as points, lines, circles and squares. Using grouping, copy, and alignment it is possible to create figures that re-use certain elements of the figure to ensure consistency and regularity. However, the true underlying constraints and regularities are often lost because the full generative history is not represented. As a consequence, it is often difficult to edit a sketch while preserving its underlying structure.

In his book "A Generative Theory of Shape" \cite{Leyton01Generative}, Michael Leyton proposes a graphics language that is totally generative and captures what he calls the maximisation of transfer and recoverability. The key idea is to describe the emergence of shape as a generative process that unfolds structure from previously unfolded substructures --- eventually going back to a single point: the origin. The maximisation of transfer means that as far as possible the shape is "explained" by re-using existing building blocks. Once a given shape is understood in terms of such a totally generative history, it can be intelligently manipulated by changing sub-structures (which may appear repeatedly in the unfolded shape), completing an incomplete shape based on the inferred regularities or using it as a building block in a super-structure.


To make Leyton's theory practical, we introduce the stochastic wreath process, which generalizes Leyton's formalism to the case of noisy shapes. While Leyton's generative theory of shape characterises a given highly regular shape, the stochastic wreath process represents a distribution over shapes---which have irregular appearance but highly regular structure. The noise process factorizes across the different hierarchical levels of the shape (one per group factor in the chain of wreath products), and hence is perfectly aligned with the generative process.

The stochastic wreath process allows us to make Leyton's theory practical in the sense that for a given hand-drawn sketch of a shape we can infer a posterior distribution over generative histories in terms of the wreath process. To explore this idea, we define a rendering pipeline based on the wreath process which generates actual pixel images and propose a reversible jump MCMC method \cite{Green95reversiblejump} inspired by Approximate Bayesian Computation \cite{Wilkinson08} for inference. Note that the model class we describe can also be viewed as a domain specific probabilistic programming language with the inference process attempting to synthesize appropriate models.

\pdfoutput=1
\section{A GENERATIVE MODEL OF SHAPE}
\label{sec:shape}

\subsection{ LEYTON'S GENERATIVE THEORY} 
Leyton \cite{Leyton01Generative} characterizes the structure of a shape
by the (ordered) sequence of actions on the canvas that led to its creation, its generative history. In a broad sense, geometric objects are seen as memory stores of a set of actions. These actions are modelled by a series of (algebraic) groups of transformations. 
\subsubsection{Preliminaries}
\begin{definition} \emph{(Groups)} A group $(G, \circ)$ is a nonempty set G together with a binary operation $\circ$ on $G$ that satisfies the following properties:
\[\begin{array}{ll}
	   \text{(Closure)} & \forall g_1, g_2 \in G, g_1 \circ g_2 \in G
	\\ \text{(Associativity)} & g_1 \circ (g_2 \circ g_3) = (g_1 \circ g_2) \circ g_3, \\ &\forall g_1, g_2, g_3 \in G
	\\ \text{(Neutral element)} & \exists e \in G, \text{ s. t. } e \circ g = g \circ e = g, \forall g \in G
	\\ \text{(Inverse)} & \forall g \in G, \exists h \in G, \text{ s.t. } g \circ h = h \circ g = e \\
	& \text{ and } h \text{ will be denoted by } g^{-1}.
\end{array}\]
\end{definition}
\begin{definition} \emph{(Group action)}
Let $G$ be a group and $X$ be a set. $G$ is said to act on $X$ if there is a map $\alpha$ such that: $\alpha(e, x) = x, \forall x \in X$ where $e$  is the neutral element of  $G$ and that $\alpha(g, \alpha(h,x)) = \alpha(g \circ h, x), \forall g, h \in G$.
\\Commonly the abbreviation $\alpha (g,x) = gx$ is used.
\end{definition}
In this paper we consider three major families of transformation groups.
These transformations act on a two dimensional vector space $\mathbb{R}^2$.
To simplify the presentation, we work within an extended space with elements in $\mathbb{R}^3$, where the third dimension is set to unity such that affine transformations can be expressed as matrix multiplications.
\begin{itemize}
	\item Translations (along the $X$ and $Y$ axis):
	\begin{equation}
				G_T = \{T_t | t \in \mathcal{I} \} \cong \{A_{T_t} | t \in \mathcal{I} \}
		\label{eq:def_GT}
	\end{equation}	
	where 
	\begin{equation*}
			A_{T_t} =
			\left( \begin{array}{cc|c}
			1 & 0 & t\\
			0 & 1 & 0\\ \hline
			0 & 0 & 1\end{array} \right) 
	\end{equation*}
	for translations along the $X$-axis. And which can be either continuous ($\mathcal{I} = \mathbb{R}$) or discrete ($\mathcal{I} = \mathbb{Z}$).
					
	\item Rotations (about the origin $O$) of discretization $n$:
	\begin{equation}
		G_R = \{R_k| k \in \mathcal{I}\} \cong \{A_{R_k}| k \in \mathcal{I}\}
		\label{eq:def_GR}
	\end{equation}
	where $n = |\mathcal{I}|$,
		\begin{equation*}
			A_{R_k} =
		\left( \begin{array}{cc|c}
		\cos{\theta} & \sin{\theta} & 0\\
		- \sin{\theta} & \cos{\theta} & 0\\ \hline
		0  & 0 & 1\end{array} \right) \text{ with } \theta = \frac{2\pi k}{n}
	\end{equation*}
	and which can be infinite ($\mathcal{I} = \mathbb{Z}_n$ with $n  \rightarrow \infty$), or discrete ($\mathcal{I} = \mathbb{Z}_n$, with $n \in \mathbb{N}$, $n\geq 2$, finite), where $\mathbb{Z}_n$ denotes the cyclic group of order $n$. 

	\item Mirroring/Reflexion (along the $X$ and $Y$ axis):
	\begin{equation}
		G_M = \{M_k| k \in \mathcal{I}\} \cong \{A_{M_k}| k \in \mathcal{I}\}
	\end{equation}
	where
		\begin{equation*}
			A_{M_k} =
		\left( \begin{array}{cc|c}
		1 & 0 & 0\\
		0 & (-1)^{k} & 0\\ \hline
		0  & 0 & 1\end{array} \right)
	\end{equation*}
	for mirroring along the $X$-axis, and $k \in \mathcal{I} = \mathbb{Z}_2$.
\end{itemize}


\subsubsection{Generating/Drawing a line}
Let us consider the scenario of drawing a horizontal line: we start with a point, call it $p$, and in order to create a line, we translate this point along the horizontal direction. We will model this action by a continuous translation group as defined in Eq. \ref{eq:def_GT}. More precisely, for each element in translation-group $T_t$ we make a copy of the point $p$, call it $p_t$, and then transfer $p_t$ to the desired location by letting $T_t$ act on it: this will produce a new point on the canvas at $t$ distance away from $p$.
\par If we were to apply the whole (continuous) translation group $G_T(\mathcal{I})$, where $\mathcal{I} = \mathbb{R}$, we obtain an infinite line around the starting point $p$. This is not always desirable, instead more commonly we would like to draw a (bounded) line-segment. This can be done by noticing that in order to create a segment we need only a subset of the elements of the full translation group considered above. We will call the subset of indices of these group elements (that are presented in the picture) the occupancy set, $occ \subset \mathcal{I}$, associated with the desired shape. In the case of continuous groups, we allow $occ$ to be specified as an interval (i.e. the unit segment centred around the origin will have occupancy $[-0.5,0.5]$) and for the discrete cases, $occ$ can be an arbitrary selection of the available indices in $\mathcal{I}$. Furthermore, we will often use the following terminology to describe the occupancy set: \textit{full} occupancy if $occ = \mathcal{I}$; \textit{single} occupancy if $|occ| = 1$ (set contains only $1$ element) or \textit{arbitrary} otherwise.

\subsubsection{Generating/Drawing a square}
\begin{figure}[ht]
\begin{center}
\centerline{\includegraphics[width=1\columnwidth]{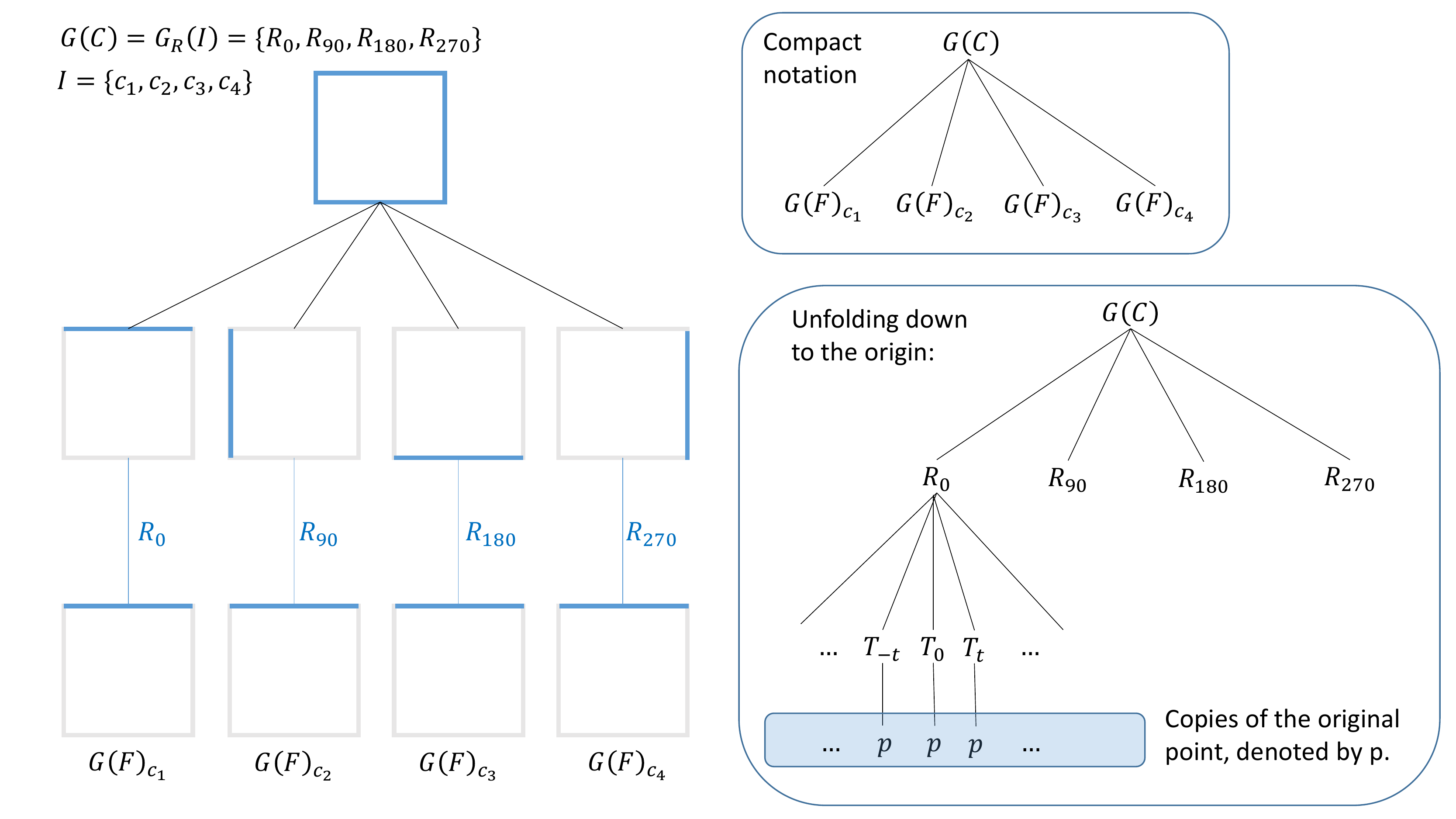}}
\caption{Wreath Product: Generation of a square}
\label{fig:TreeStruct}
\end{center}
\vskip -0.1in
\end{figure}

Let us move on to more complex structures and consider describing the generative process behind drawing a square. Remember that the central idea of this generative process is maximization of transfer. We have previously seen how to characterize the generative process of drawing a line-segment, which could be one side of the square. Let us start by drawing the top side of the square. To create a square we wish to transfer this side via a $4$-fold rotation group (Eq. \ref{eq:def_GR}). As before, we start by creating copies of the top side, one for each element in the transformation group for a total of four. Then we let each element of the group act on its copy to create each side of the square. The process is depicted in Figure \ref{fig:TreeStruct}.
\par The generative process of the square, as described so far, starts with any point $p$ on the top side or --- given the right choice of occupancy --- indeed anywhere on the implied infinite line of which the side is a finite segment. In this paper, all the generative processes we are considering will start at the origin (defined as the centre point of the canvas). Thus, the complete generative history of a square will start by translating the origin to a point $p$ on the top side of the square. This translation will have an occupancy set of cardinality $1$, containing only the index corresponding to the translation transformation that maps the origin onto $p$ --- it will leave no trace. After this initial translation of the origin, the generative process continues as described above.

\subsubsection{Transfer as a Wreath product}
 In both scenarios what happens amounts to two steps: we make copies of the original shape, one for each element in the transformation group and then for each of these copies we apply the corresponding group element to transfer it to its desired location, form, or orientation. This construction is algebraically modelled by the wreath product between the initial shape generative history and the transformation group we want to employ in the transfer.

\begin{definition} \emph{(Semidirect products)}
Consider $H$ and $N$ two groups, with their respective group operations $\circ_{H}$ and $\circ_{N}$ and a group homomorphism $\tau: H \rightarrow Aut[N]$ \footnote{$Aut[N]$ is the automorphism group of $N$. An automorphism of an object is a isomorphic map from the object to itself - i.e a mapping from the object to itself preserves the structure. Additionally, it can be shown that the set of all automorphisms of an object forms a group under the composition operation.}. Now, let $G$ be the set of ordered pairs $\langle h,n \rangle$, with $h \in H$, $n \in N$. We can define a binary operation $\circ_{G}$ on $G$ as follows:
\begin{equation*}
	 \langle n_1,h_1 \rangle \circ_{G} \langle n_2, h_2 \rangle = \langle \tau(h_2)[n_1] \circ_{N} n_2, h_1 \circ_{H} h_2\rangle
\end{equation*}
for all $ \langle n_1,h_1 \rangle,  \langle n_2,h_2 \rangle \in G$. Then under this operation, $G$ is a group, denoted by $N \circledS_{\tau} H$ and referred to as the \emph{semidirect product of N and H under $\tau$}.
\end{definition}

\begin{definition} \emph{(Wreath product)}
Let $A$ and $H$ be two groups and let $\Omega$ be a set with $H$ acting on it. Let $A^{\Omega}$ be the direct product of the copies of $A^{\Omega} := A$ indexed by the set $\Omega$: $A^{\Omega} = \prod_{\omega \in \Omega}{A_{\omega}}$. Then we can define the action of $H$ on $A^{\Omega}$ in a natural way by letting the group action of $H$ act on the indices of the product:
\begin{equation*}
\beta_h : \begin{cases}
	\prod_{\omega \in \Omega}{A_{\omega}}  &\rightarrow \prod_{\omega \in \Omega}{A_{h	\omega}}  \\
	(a_{\omega_1}, a_{\omega_2}, \cdots )  &\rightarrow (a_{h\omega_1}, a_{h\omega_2}, \cdots )
\end{cases}
\end{equation*}
Given this action, the wreath product of $A$ and $H$ is defined as the semiproduct $[A^{\Omega}] \circledS_{\tau} H$, where $\tau$ is implicitly given by the action above,  $\tau: h \longmapsto \beta_h$.
\end{definition}

We can see from this definition that in order to define a wreath product we need two groups and a set. The first one, $A$, will correspond to the generative history of the initial shape $S_0$ that we want to transfer and the second one, $H$, will correspond to the transformation group used in the transfer.\footnote{ Terminology used in Leyton's literature: the generative history of the initial shape $S_0$ that we would like to transfer is referred to as the fibre-group and will be denoted by $G(F)$; the group responsible for the transfer of $S_0$'s generative history will be referred to as the control-group and will be denoted by $G(C)$. } And the set $\Omega$ corresponds to the set of indices $\mathcal{I}$ associated with the transformation group. At this point, it should be clear that the descriptions of the generative structures will be groups under full occupancy. Partial occupancy, i.e., the situation in which only a subset of elements of the shape is displayed, can be integrated into the group-theoretical framework by appending cyclic $Z_2$ switches to the fibre copies before application of the control group. These can be seen as colour or on/off switches for parts of the shape.

\subsubsection{Shapes as n-fold wreath products}
We would like to apply the same concept of transfer several times in the formation of an image in order to maximize re-usability.
For instance consider that once we have a square we would like to form a circle of four such squares.
Using the ideas highlighted above, we could transfer the already constructed square by another 4-fold rotation group. But first, recall that the rotations we consider are only around the origin and the previously formed square is centered about the origin - thus if we were to apply a 4-fold rotation to this square, we will end up with four coinciding copies of this square. Therefore, first we need to translate the square by the intended radius of the circle we want to form.
 This will give rise to a new shape that can be characterized by a 5-fold wreath product:
\begin{equation*}
	G_0 \text{\textcircled{w}} G_1 \text{\textcircled{w}} G_2 \text{\textcircled{w}} G_3 \text{\textcircled{w}} G_4 \text{\textcircled{w}} G_5
\end{equation*}
where 
\begin{itemize}
	\item $G_0 = \{e\}$ is the trivial group corresponding to the origin (it transfers the origin onto itself)
	\item $G_1$ is the continuous translation group $G_1 \cong \mathbb{R}$, responsible for the vertical translation of the origin to the point $p$ on the top side.
	\\$occ_1 = {0.5} \subset \mathbb{R} (=\mathcal{I}_{G_1})$
	\item $G_2$ is the continuous translation group $G_2 \cong \mathbb{R}$, responsible for producing an infinite line.
	\\$occ_2 = [-0.5,0.5] \subset \mathbb{R} (=\mathcal{I}_{G_2})$
	\item $G_3$ is the first 4-fold rotation group  $G_3 \cong \mathbb{Z}_4$, responsible for producing a square.
	\\$occ_3 = \{0,1,2,3\} = \mathbb{Z}_4 (=\mathcal{I}_{G_3})$
	\item $G_4$ is the discrete translation group $G_4 \cong \mathbb{Z}$, responsible for translating the formed square $2$ units away from the origin (anticipating the rotation)
	\\$occ_4 = \{2\} \in \mathbb{Z} (=\mathcal{I}_{G_4})$
	\item $G_5$ is the second 4-fold rotation group $G_5 \cong \mathbb{Z}_4$, responsible for producing the circle of squares.
	\\$occ_5 = \{0,1,2,3\} = \mathbb{Z}_4 (=\mathcal{I}_{G_5})$
\end{itemize}

\subsubsection{A Grammar for Shapes}
While Leyton develops his theory as abstract mathematics, our concern is to have a concrete representation suitable for probabilistic inference.
Hence, we introduce the following grammar to represent shapes.
\begin{eqnarray*}
\mathcal{S} &::=& [(G_1, occ_1); \dots; (G_n, occ_n)] \\
G &::=& \ttranslate\ \taxis \mid \trotate\ m \mid \tmirror \\
\taxis &::=& \tx \mid \ty \\
occ &::=& [k_1; \dots; k_n] \mid [r_1,r_2]
\end{eqnarray*}

A shape $\mathcal{S}$ denotes (1) the generative history, the $n$-fold wreath product
$G_0 \text{\textcircled{w}} G_1 \text{\textcircled{w}}\cdots  \text{\textcircled{w}} G_n$
where $G_0=\{e\}$ is the implicit trivial group corresponding to the origin,
but also (2) the associated occupancies $occ_i$ for each level, that characterize the elements of this structure observed in the picture.

We write our circle of squares example as follows:
\[ [ \begin{array}[t]{@{}l}
     (\ttranslate\ \ty, [0.5,0.5]), \\
     (\ttranslate\ \tx, [-0.5,0.5]),
     (\trotate\ 4, [0..3]), \\
     (\ttranslate\ \tx, [2]),
     (\trotate\ 4, [0..3])]
  \end{array} \]

\subsection{THE WREATH PROCESS: STOCHASTIC SHAPE}
\begin{figure}[ht]
\vskip 0.2in
\begin{center}
\centerline{\includegraphics[width=1\columnwidth]{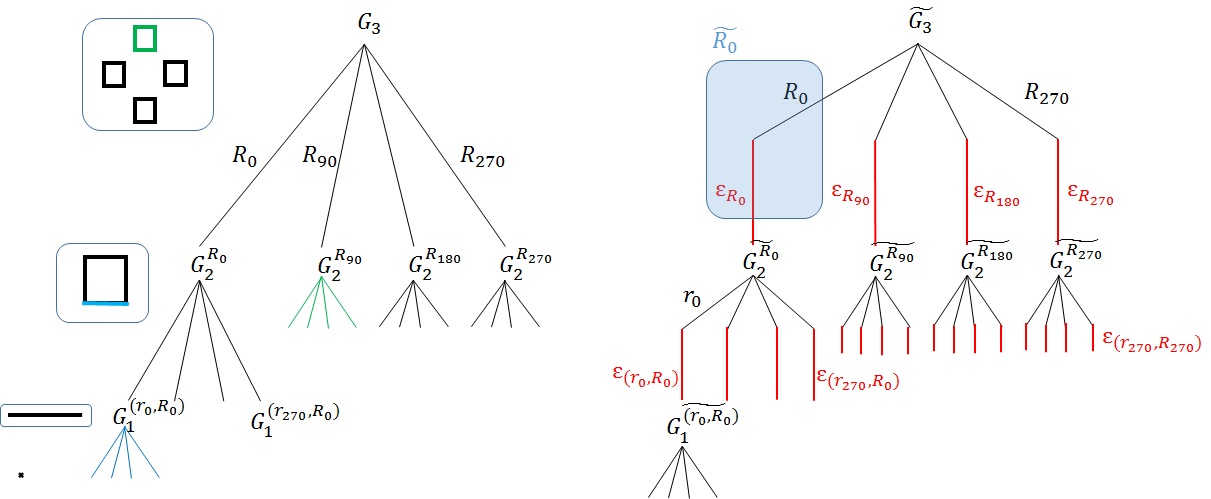}}
\caption{Noise model. For compactness, we omit the single occupancy levels in the generative history: but there are part of the structure and will take noise instances as well}
\label{fig:NoiseTree}
\end{center}
\vskip -0.2in
\end{figure} 

To model hand-drawn sketches, we define a noise process that accounts for the imperfections present in free-hand drawings, and that arises naturally by perturbing the generative history of the intended shape.
More precisely, each transformation present in the generative history of our shape will have a noise level $\epsilon$ which accounts for the error made by the user when trying to perform the transfer corresponding to that transformation.

Let us consider that the intended structure (shape) is given by an $n$-fold wreath product
($ G_0 \text{\textcircled{w}} G_1 \text{\textcircled{w}} \cdots  \text{\textcircled{w}} G_n$)
then $\forall g \in G_i, \forall i$, a level of noise will be sampled to account for each application of each element in the control-group.
In the generative history an element of a group $G_i$ is applied multiple times, as many times as it is copied by the levels above it.
Although in the exact shape the copies of these transformations are the same,
under the noise process each of these copies receives its own perturbation independent of the noise instances corresponding to other noisy copies:
$\forall g^{(j)} \in G^{(j)}_i$ sample $\epsilon^{(j)} \sim p_{G_i}(\epsilon)$, where $j \in occ_{i+1}\times \cdots occ_n$
and can thought of as the coordinate of the subtree on which $g^{(j)}$ we act, out of the copies of that fibre-group that were created by repeat transfer (each time a new transformation group was applied).
Defining the probability distribution $p_{G}(\epsilon)$ of possible perturbations for a group $G$ we assume that there exits an embedding of $G$ into a bigger (continuous) group, on which both noisy and non-noisy transformations are defined. \footnote{Usually there is a natural way of defining this embedding, but we could also construct an embedding by another wreath product, if there is no other more trivial embedding.} 
	\begin{table}[H]
		\caption{Samples of the wreath process for the square, four-fold rotation of the square}
	\begin{tabular}{p{0.9cm} p{0.9cm} p{0.9cm} p{0.9cm} p{0.9cm} p{0.9cm}}
 \\ [-0.2cm] 	\hline \\ [-0.2cm]
	\fbox{\includegraphics[width=0.12\columnwidth]{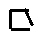}}  & 
	\fbox{\includegraphics[width=0.12\columnwidth]{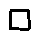}}  &
	\fbox{\includegraphics[width=0.12\columnwidth]{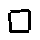}}  &
	\fbox{\includegraphics[width=0.12\columnwidth]{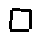}}  &
	\fbox{\includegraphics[width=0.12\columnwidth]{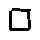}}  &
	\fbox{\includegraphics[width=0.12\columnwidth]{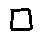}} \\ [-0.2cm] \\\hline \\ [-0.2cm]
	\fbox{\includegraphics[width=0.12\columnwidth]{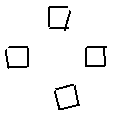}}  & 
	\fbox{\includegraphics[width=0.12\columnwidth]{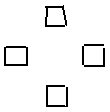}}  &
	\fbox{\includegraphics[width=0.12\columnwidth]{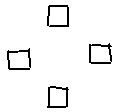}}  &
	\fbox{\includegraphics[width=0.12\columnwidth]{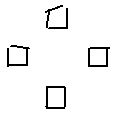}}  &
	\fbox{\includegraphics[width=0.12\columnwidth]{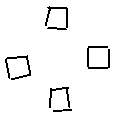}}  &
	\fbox{\includegraphics[width=0.12\columnwidth]{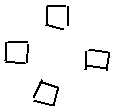}} 
 \\ \\[-0.2cm] \hline
		\end{tabular}
	\label{tab: noisy_square}
	\vskip -0.2in
	\end{table}
	
For us, this is trivially the case as we have already considered this embedding when defining the continuous occupancy. Both the continuous and discrete versions of the translation and rotation groups can be embedded in the continuous groups.
The noisy transformations obtained by composing $g$ with the sampled noise instances are applied to the corresponding fibre-copies.
We denote the set formed by these noise actions on the 2D plane as $\tilde{G_i} = \{\tilde{g}_k^{(i)}| \forall g_k^{(i)} \in G_i\}$ for each control group $G_i$.
The process is illustrated in Figure \ref{fig:NoiseTree}. Under this interpretation we can define a noisy shape as $	\tilde{\mathcal{S}} \sim W(\cdot|\mathcal{S})$.
In the following we will denote by $\mathcal{N}$ the set of all noise instances corresponding to a shape $\mathcal{S}$.
\subsection{PRIOR OVER HAND-DRAWN SHAPES}
Under the Bayesian paradigm, we need to specify a prior over the model's parameters $(\mathcal{S}, \mathcal{N})$: 
\par \textbf{Prior over generative histories $p(\mathcal{S})$}
	\begin{eqnarray}
	p(\mathcal{S}) &=& p([(G_1, occ_1); \dots; (G_n, occ_n)]) \nonumber \\
	&=& \prod_{i=1}^{n}{p(occ_i|G_i)p(G_i)}
	\end{eqnarray}
	where, in the our particular case of transformation groups: $p(G_i)$ is considered to be uniform at any level $i$. The prior over $occ_i$ reflects a strong preference for \textit{single} (with probability $p$) or \textit{full} occupancy (with probability $p$) and with $(1-2p)$ probability we will consider \textit{special} occupancy, which will be sampled as follows: if $|\mathcal{I}| < \infty$ then a particular element of the index set is to be switched on with probability $\frac{1}{|\mathcal{I}|}$, otherwise we define a bounding parameter $B$ (sampled around the origin) which restricts $|\mathcal{I}|$ to a finite set and then we will sample uniformly within that restriction. For continuous occupancy, in this work, we will restrict ourselves to line-segments of unit length and full circles. 
	
\par \textbf{Noise instances prior $p(\mathcal{N}|\mathcal{S})$}:
	\begin{eqnarray}
	p(\mathcal{N}|\mathcal{S}) &=& \prod_{i=1}^{n} p({\overrightarrow{\epsilon}_i}|\theta_{\mathcal{N}}, \mathcal{S})\nonumber \\ & = & \prod_{i=1}^{n} {\prod_{j \in occ_i\times\cdots\times occ_n}{p(\epsilon_i^{(j)}|\theta_{\mathcal{N}}, G_i)}}
	\end{eqnarray}
 where $\theta_{\mathcal{N}}$ are the hyper-parameters governing the distribution of the noise instances and $\overrightarrow{\epsilon}_i$ is the vector of noise instances at level $i$. This vector has $|occ_i| \times \cdots \times |occ_{n}|$ elements. The prior over the noise instances is as follows: for translation  $\epsilon_{trans} \sim \mathcal{N}(0, \sigma_{trans}) $ where $\sigma_{trans} \sim \Gamma(1, 20)$; for rotation: $p(\epsilon_{Rot}) = \left(\frac{e^{\cos(\epsilon_{Rot})/\sigma_{Rot}^2}}{2\pi I_0(\sigma_{Rot}^{-2})} \right) $ with $\sigma_{Rot} \sim \Gamma(\pi/n, n^2)$, where $I_0(k)$ is the Bessel function of order $0$ ($I_0(k) = \frac{1}{\pi} \int_{0}^{\pi}{e^{k \cos{x}} dx}$); for mirroring: we do not consider an explicit noise action, but by the way we defined the wreath process, a mirrored copy of a given noisy shape $\tilde{S}_0$ will be different from $\tilde{S}_0$, as this noisy copy of $S_0$ will have its own noise instances---sampled independently of the noise instances of $\tilde{S}_0$. 

\section{INFERENCE}
\label{sec:inference}
Consider the grey-value mapping $I_D$ of an image.
Our aim is infer the generative history of the shape in this image.
This amounts to inferring a structure $\mathcal{S}$ representing the $n$-fold wreath product describing the underlying symmetries present in the shape, plus their observed occupancy. 
Using Bayes' rule, we can express the posterior probability as:
\begin{equation}
	p(\mathcal{S}| I_{D}) \propto p(I_D| \mathcal{S} ) p(\mathcal{S})
\end{equation}
Thus, for this computation, we need to specify and evaluate a prior over generative histories of shapes and a likelihood that evaluate the input data on a given shape $\mathcal{S}$. The prior encodes our beliefs about the kind of shapes we expect to see and may be chosen conveniently to ensure tractability. That leaves us with the computation of the likelihood, which in this case is non-trivial as $\mathcal{S}$ and $I_D$ occupy very different domains. However, this fairly common problem that can be addressed  using approximate Bayesian computation (ABC) methods that bypass the (exact) evaluation of the true likelihood. The idea is as follows: given a parameter setting $\theta$, a dataset $\tilde{D}$ can be simulated for the stochastic model specified by $\theta$. Then a distance measure, $\rho$, can be defined between the input data $D$ and $\tilde{D}$ (that now live in the same space). If the simulated data does not match the input data within a given tolerance, i.e. $\rho(\tilde{D}, D) > \epsilon$, then the set of parameters will be rejected. This idea was particularized to MCMC simulations in \cite{Marjoram03} and \cite{Sisson07}. Note that these require the specification of the threshold $\epsilon$, which might require further inference. However, more recently in \cite{Wilkinson08} and \cite{Mansinghka2013}, it was shown that the hard specification of this threshold could be replaced by a stochastic likelihood model. Our inference will combine these ideas, as described in the next section.

\subsection{ABC FOR THE GENERATIVE HISTORY OF SHAPES}
We start by rendering the image corresponding to $\mathcal{S}$ to get into the same domain as the data $I_D$ using a deterministic rendering function $f(\mathcal{S}) = I_{\mathcal{S}}$.
In principle, we could define a measure on the space of images and compute how far the exact rendering of our proposed model is with respect to the input image $I_D$. But defining such a measure is problematic, as most such measures will induce very sharp distributions on the rendered image $I_R$. The likelihood will yield high values for exact and almost exact matches, while most other models will be given a likelihood close to zero. In other words, this approach will likely fail to discriminate close solutions from arbitrary proposals, except for the unlikely case that an (almost) perfect match is rendered.
\par To overcome this problem, we make use of the ideas in approximate Bayesian computation highlighted before and follow the approach in \cite{Mansinghka2013}.  We estimate the likelihood $P(I_D|\mathcal{S})$ using a stochastic likelihood model based on a stochastic image renderer. Instead of rendering the exact image, we render a noisy version of it as illustrated in Tables~\ref{tab: noisy_square}. In addition, to increase stochasticity we apply a Gaussian blur to the rendering, specified by $\mathcal{X} = \{w_b, \sigma_b\}$.
\begin{equation}
	\mathcal{S} \xrightarrow{\mathcal{N}} \tilde{I}_{D} \xrightarrow{\mathcal{X}} I_R
\end{equation}
where $\tilde{I}_{D}$ is generated via the wreath process, given $\mathcal{S}$ and $I_R$ is the image resulting from applying a Gaussian blur of $\sigma_b$ and a window size of $2w_b$.
Under this formulation and keeping in mind that the rendering function although stochastic by nature, becomes deterministic given $\mathcal{N}$ and $\mathcal{X}$, the posterior probability can be approximated as:
\begin{eqnarray*}
	p(\mathcal{S}|I_D) &\propto & \int p(\mathcal{S})\underbrace{p(I_R|\mathcal{S},\mathcal{N}, \mathcal{X})}_{\delta_{f(\mathcal{S},\mathcal{N},\mathcal{X})}(I_R)}p(\mathcal{N},\mathcal{X}|\mathcal{S})p(I_D|I_R) \\
	&\propto & \int p(\mathcal{S}) \delta_{f(\mathcal{S},\mathcal{N},\mathcal{X})}(I_R) p(\mathcal{N}|\mathcal{S})p(\mathcal{X})p(I_D|I_R)
\end{eqnarray*}

Prior over the parameters of the Gaussian blur $p(\mathcal{X})$: $w_b \sim b_{w}\cdot Beta(1,2)$ and $\sigma_b \sim b_{\sigma} \cdot \Gamma(1,1)$.

\subsection{EMPIRICAL LIKELIHOOD} 
The stochastic render will produce a noisy instance of model $I_R$ against which the input image $I_D$ can be evaluated. To this end, we define the empirical likelihood, assuming a Bernoulli distribution in pixel space:
\begin{equation*}
	p(I_D|I_R) = \prod_{[x,y]} {I_R[x,y]^{I_D[x,y]} (1-I_R[x,y])^{(1-I_D[x,y])}}
\end{equation*}
where $I_R[x,y] \in [0,1]$ denotes the (grey-scale) intensity of pixel located at $[x,y]$ in image $I_R$ and is interpreted here as the probability of that pixel being black.

\subsection{REVERSIBLE JUMP-MCMC FOR THE WREATH PROCESS} \label{sec:RJMCMC}
In this section, we will give a general-purpose algorithm for inference in wreath processes, based on Reversible Jump-MCMC (introduced in \cite{Green95reversiblejump}, and refined in \cite{Green09reversiblejump}), but particularize it in the proposals to exploit the structure of the wreath product. The idea is the following: we assume that the upper-level structure (of the top level groups) has the greatest impact on the appearance of a shape and hence should be kept more stable than lower-level parts of the generative history. In other words, lower levels can be explored given upper levels, but changes in upper levels are likely going to lead to major revisions in the lower levels. Assuming that the higher-level structure has been detected, we propose objects on which this upper structure acts by transfer. 
\begin{algorithm}
\caption{Reversible-Jump Markov Chain Monte Carlo}
\label{alg:RJMCMC}
\begin{algorithmic}
	 \STATE (RJ.0) \textit{Start with random model $(k_0, \theta_{k_0})$, of non-zero probability. }
	 \LOOP
	 \STATE (RJ.1) \textit{Propose a {visit} to model $m_{k'}$ with probability $j(k\rightarrow k')$} \\[0.2cm]
	 \STATE (RJ.2) \textit{Sample $u \sim q(u| \theta_k, k, k')$(proposal density)} \\[0.2cm]
	 \STATE (RJ.3) \textit{Let the new sample be defined by $(\theta_{k'}, u') = g_{(k, k')}(\theta_{k}, u)$, where  $g_{(k, k')}$ is bijective with its inverse being $g_{(k', k)}$ and random variables $u$ and $u'$ play the role of matching the dimensionality of the embeddings $(\theta_{k'}, u')$ and $(\theta_{k}, u)$: i.e.
	 $dim(\theta_{k'})+ dim(u') = dim(\theta_k)+dim(u)$} \\[0.2cm]
	 \STATE (RJ.4) \textit{Accept new model $(k',\theta_{k'})$ with probability $\alpha(\theta_k \rightarrow \theta_{k'}) = \min{\left( 1, A_{\theta_k \rightarrow \theta_{k'}} \right) }$, where:
	\begin{eqnarray}
		A_{\theta_k \rightarrow \theta_{k'}} &=& \frac{\pi(\theta_{k'}, k')}{\pi(\theta_k, k)} \frac{j(k'\rightarrow k)}{j(k\rightarrow k')} \frac{q(u'| \theta_{k'}, k', k)}{q(u| \theta_k, k, k')} \nonumber \\
		&&\left| {\frac{\partial g_{(k, k')}(\theta_k, u)}{(\theta_k, u)}}\right| 
	\label{eq:alpha_JR}	
	\end{eqnarray}} 
	 \ENDLOOP
\end{algorithmic}
\end{algorithm}
Let us look at what Algorithm \ref{alg:RJMCMC} amounts to in this case. Given a model $\theta_n = (\mathcal{S}, \mathcal{N}, \mathcal{X})$ where $n$ indicates that $\mathcal{S}$ is an $n$-fold wreath product, we wish to propose a new model $\theta_{n'} = (\mathcal{S'}, \mathcal{N'}, \mathcal{X'})$. First with a given probability we choose which parameters to resample: $\mathcal{S}$, $\mathcal{N}$ or $\mathcal{X}$. Varying $\mathcal{S}$, in most cases varies $\mathcal{N}$ too. The other two types of parameters can be sampled independently and when possible we would like to keep all other parameters fixed---we are making local changes only in one dimension type at a time. Thus in our case:
	\begin{eqnarray}
	 q((k,\theta_{k}) \rightarrow (k',\theta_{k'})) &=&  q(\mathcal{S} \rightarrow \mathcal{S}') q(\mathcal{N} \rightarrow \mathcal{N}'|\mathcal{S}') \nonumber \\
	 & &\cdot q(\mathcal{X} \rightarrow \mathcal{X}')q(\lambda \rightarrow \lambda') q(k \rightarrow k') \nonumber 
	\end{eqnarray}
where $\lambda$ is a global scaling factor, that is sampled uniformly between $[1,50]$. This corresponds to our (inferred) unit. Freehand sketches might not respect a standardized unit (like $1$cm), but we postulate that the user has in mind an implicit grid with this $\lambda$ as unit interval. 
\par \textbf{Noise instances proposal:} $q(\mathcal{N} \rightarrow \mathcal{N}') = p(\mathcal{N}'|\mathcal{S})$
\par \textbf{Gaussian blur proposal:} $q(\mathcal{X} \rightarrow \mathcal{X}') =p(\mathcal{X}') $
are sampled from the prior, which was previously described. Let us concern ourselves with the proposals in the structure of the wreath product:
\par \textbf{Shape proposal:} $q(\mathcal{S} \rightarrow \mathcal{S}')$
Given $\mathcal{S} = [(G_1, occ_1); \dots; (G_n, occ_n)]$, we propose a new structure and implicitly a new shape as follows. We have two main types of moves: one that changes the dimensionality of the structure through $n$, and one that does not. The moves within a model (keeping $n$ constant) will change the occupancy sets or the individual groups, but will prefer to move within the same family of transformation groups. Changes in the dimensionality of the wreath product are as follows: we pick a random level $i$ (with higher probability for lower $i$-s), at which to segment the structure. We keep the upper-level structure from level $i+1$ to $n$ and we re-sample (from the prior) the group on which this structure acts. As mentioned before, this corresponds to keeping the higher-level symmetries and changing the object on which these symmetries act. 
The nature of these proposals keeps a nested structure over models, that with sampling components from the prior greatly simplifies the computation of the acceptance ratio in Eq. \ref{eq:alpha_JR}, as the determinant factor is always $1$ and several other simplifications are possible as the prior factorizes over levels and new fibre groups are sampled from this prior.	
	\begin{table}[htc]
	\label{tab:proposals} 
		\caption{One sample run, illustrating various types of proposals: (a)$\rightarrow$(b) Change in occupancy of the top level translation; (b)$\rightarrow$(c) Changing the fibre-group(from a circle to a cross), but keeping the top level structure; (c)$\rightarrow$(d) Another change in the fibre-group, as the previous rotation did not quite fit the data; (d)$\rightarrow$(e) Change in the blur parameters: once the structure(at least up to occupancy) is inferred; (e)$\rightarrow$(f) Change in the noise instances.}
	\begin{tabular}{p{0.7cm}| p{0.75cm} p{0.75cm} p{0.75cm} p{0.75cm} p{0.75cm}  p{0.75cm}}
	\hline
	\multicolumn{1}{ c| }{Input} & \multicolumn{6}{ |c }{Intermediate iterations} \\ \hline
	& (a) & (b) & (c) & (d) & (e)& (f) \\ 
	\fbox{\includegraphics[width=0.08\columnwidth]{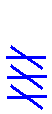}} &
	\fbox{\includegraphics[width=0.09\columnwidth]{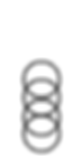}}  & 
	\fbox{\includegraphics[width=0.09\columnwidth]{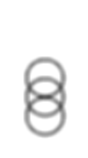}}  & 
	\fbox{\includegraphics[width=0.1\columnwidth]{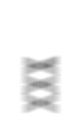}}  &
	\fbox{\includegraphics[width=0.1\columnwidth]{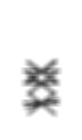}} 
&
	\fbox{\includegraphics[width=0.1\columnwidth]{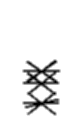}}	
	&
	\fbox{\includegraphics[width=0.1\columnwidth]{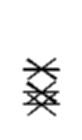}}
	\\ \\ [-0.2cm]
	&
	\fbox{\includegraphics[width=0.09\columnwidth]{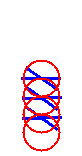}}  & 
	\fbox{\includegraphics[width=0.09\columnwidth]{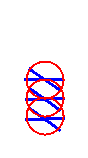}}  & 
	\fbox{\includegraphics[width=0.1\columnwidth]{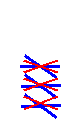}}  &
	\fbox{\includegraphics[width=0.1\columnwidth]{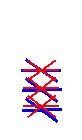}} 
&
	\fbox{\includegraphics[width=0.1\columnwidth]{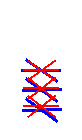}}	
	&
	\fbox{\includegraphics[width=0.1\columnwidth]{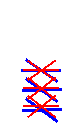}} \\ \hline \\[-0.5cm]
	\end{tabular}		
	\end{table}

\section{EXPERIMENTS}
\label{sec:experiments}
In the following, we describe a set of experiments, undertaken to illustrate the usability of the wreath process in discovering symmetry structures in noisy 2D shapes.
\par \textbf{Noise instances $\mathcal{N}$ as part of the model:} \\
Firstly, we constructed a data set of $10$ images sampled from the previously defined shape prior, which we will regard as ground truth. For each of these, we will consider: the exact rendering of the wreath product sampled and a hand-drawn version of it. On this dataset we perform two type of inferences: one as described in Section \ref{sec:RJMCMC} and one without accounting for the noise described by the Wreath process - in this case the sampling procedure is similar to the one presented in Section \ref{sec:RJMCMC}, but we do not have any proposals involving $\mathcal{N}$. This initial experiment was done to assess the impact and importance of keeping track of the noise instances $\mathcal{N}$ explicitly in the model. As a result of this experiment, we observed a substantially higher average recoverability rate of structure, especially for hand-drawn images. Thus, the rest of the experiments were carried out by inferring a model $\mathcal{M} = (\mathcal{S}, \mathcal{N})$, although in most cases the quantity we are primary interested in is $\mathcal{S}$.

\par \textbf{Measure of performance:}
\\We have defined two measures to 
quantify our results:
\begin{itemize}
	\item \textbf{(Full) Recoverability} - the inferred structure matches the group structure, or an equivalent version, including the right occupancy sets.
	\item \textbf{Recoverability up to occupancy} - the wreath product has been successfully inferred, but the occupancy is not quite right.
\end{itemize}
An example of correctly inferred structure, but with slightly off occupancy is the example in Table \ref{tab:proposals}, where the top level translation is inferred correctly, and so is the rotation group before it, but the occupancy there accounts for three elements being switched on whereas we actually observe only two. It is important to note that this is by no means optimal in assessing recoverability of structure, as for instance, the last two examples in Table \ref{tab: sampleruns} will score $0$ under both these measures, although clearly a lot of the structure, in particular the higher level control groups, are recovered. Unfortunately, quantifying partial recoverability is very problematic because of the equivalence between models under the $2D$ projection on the canvas and the limited occupancy. In general, there are several possible explanations of a partially observed structure, and various ways of constructing the same object. 

\subsection{RECOVERING SAMPLES FROM THE PRIOR}
To evaluate our model and explore its capabilities of recovering structure, we construct a data set of $50$ examples sampled from the prior. We limit the level of complexity to compositions of at most $8$ groups.{Complexities higher than that tend to lead to highly dense images, for which the number of copies of basic fiber groups tends to be very high and require a high resolution to properly distinguish them. On the other hand, if the occupancy is low, for such a dense structure, there is very little information to differentiate between possible explanations - such cases usually look random to the naked eye.} 
Sample runs can be seen in Table \ref{tab: sampleruns} and quantitative results are reported in Table \ref{tab:results_prior}.
	\begin{table}[htc]
	\vskip -0.1in
		\caption{Sample Runs}
	\begin{tabular}{p{1.4cm} |c c| c}
 \hline \\ [-0.3cm]
	\multicolumn{1}{ c| }{Input($I_D$)} & \multicolumn{2}{ |c }{Intermediate iterations} &  \multicolumn{1}{ |c }{Inferred Image(MAP)}\\ \hline
	& $I_R^{(t_1)}$ & $I_R^{(t_2)}$ & Model($I_S$) \\
	\fbox{\includegraphics[width=0.125\columnwidth]{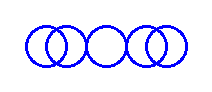}} &
	\fbox{\includegraphics[width=0.125\columnwidth]{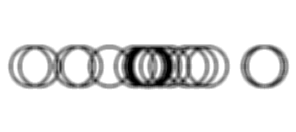}}  & 
	\fbox{\includegraphics[width=0.125\columnwidth]{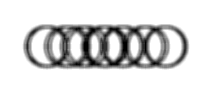}}  &
	\fbox{\includegraphics[width=0.125\columnwidth]{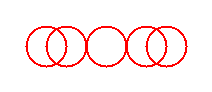}} \\
	& $I_S^{(t_1)}+I_D$ & $I_S^{(t_2)}+I_D$ & $I_S+I_D$ \\
	(Exact)&
	\fbox{\includegraphics[width=0.1\columnwidth]{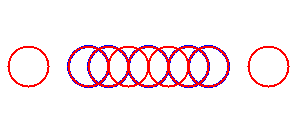}}  & 
	\fbox{\includegraphics[width=0.1\columnwidth]{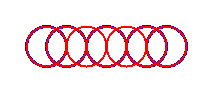}}  &
	\fbox{\includegraphics[width=0.1\columnwidth]{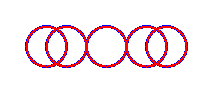}} \\
	 \hline
	\fbox{\includegraphics[width=0.1\columnwidth]{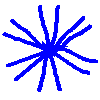}} &
	\fbox{\includegraphics[width=0.1\columnwidth]{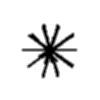}}  & 
	\fbox{\includegraphics[width=0.1\columnwidth]{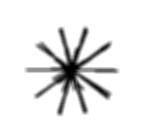}}  &
	\fbox{\includegraphics[width=0.1\columnwidth]{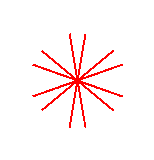}} \\
	(Drawn)&
	\fbox{\includegraphics[width=0.1\columnwidth]{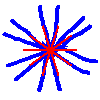}}  & 
	\fbox{\includegraphics[width=0.1\columnwidth]{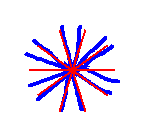}}  &
	\fbox{\includegraphics[width=0.1\columnwidth]{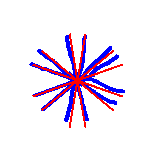}} \\
	 \hline
	 \fbox{\includegraphics[width=0.1\columnwidth]{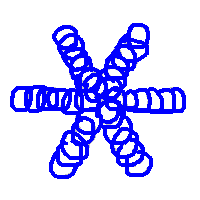}} &
	\fbox{\includegraphics[width=0.1\columnwidth]{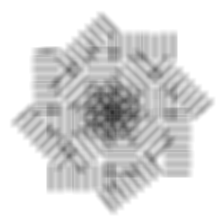}}  & 
	\fbox{\includegraphics[width=0.1\columnwidth]{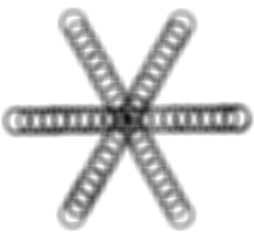}}  &
	\fbox{\includegraphics[width=0.1\columnwidth]{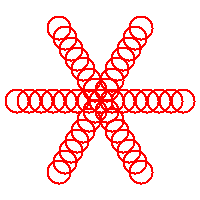}} \\
	(Drawn)&
	\fbox{\includegraphics[width=0.1\columnwidth]{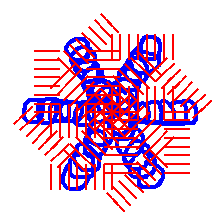}}  & 
	\fbox{\includegraphics[width=0.1\columnwidth]{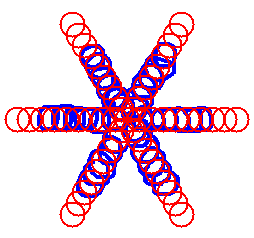}}  &
	\fbox{\includegraphics[width=0.1\columnwidth]{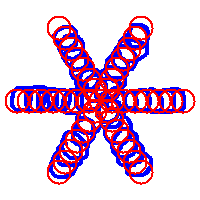}} \\
	 	\hline
	 \fbox{\includegraphics[width=0.1\columnwidth]{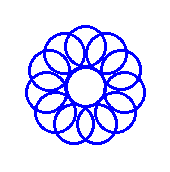}} &
	\fbox{\includegraphics[width=0.1\columnwidth]{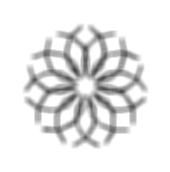}}  & 
	\fbox{\includegraphics[width=0.1\columnwidth]{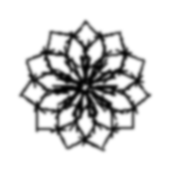}}  &
	\fbox{\includegraphics[width=0.1\columnwidth]{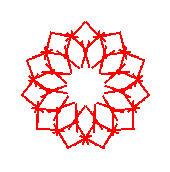}} \\
	(Exact)&
	\fbox{\includegraphics[width=0.1\columnwidth]{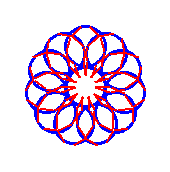}}  & 
	\fbox{\includegraphics[width=0.1\columnwidth]{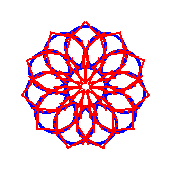}}  &
	\fbox{\includegraphics[width=0.1\columnwidth]{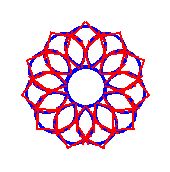}} \\
	 \hline
	 \fbox{\includegraphics[width=0.1\columnwidth]{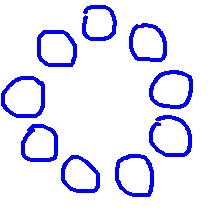}} &
	\fbox{\includegraphics[width=0.1\columnwidth]{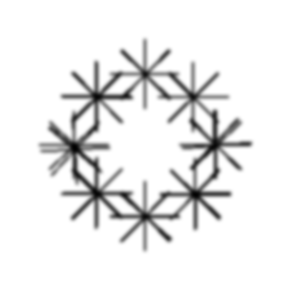}}  & 
	\fbox{\includegraphics[width=0.1\columnwidth]{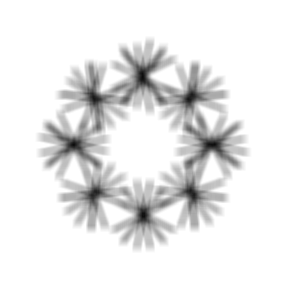}}  &
	\fbox{\includegraphics[width=0.1\columnwidth]{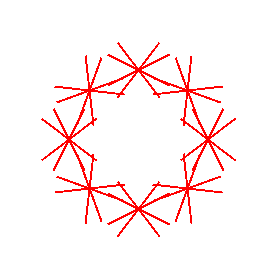}} \\
	(Drawn)&
	\fbox{\includegraphics[width=0.1\columnwidth]{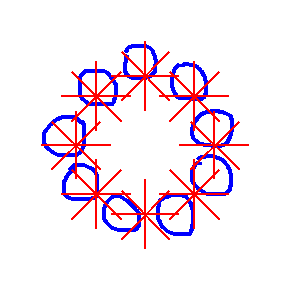}}  & 
	\fbox{\includegraphics[width=0.1\columnwidth]{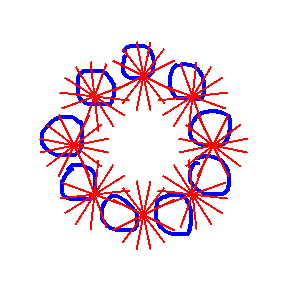}}  &
	\fbox{\includegraphics[width=0.1\columnwidth]{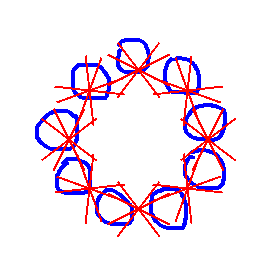}} \\
	 \hline \\[-.7cm]
	\end{tabular}
	\label{tab: sampleruns}
	\end{table}

\begin{table}[htc]
	\centering
	\label{tab:results_prior}
	\caption{Recovering structure from sample from the prior}
	\begin{tabular}{|c|c|c|}
	\hline
	 Data Set   & Recoverability & Recoverability up to occ. \\ \hline
	 Exact Shape&      $42.76 \%$ & $55.84\%$        \\ \hline
	 Hand-drawn &    $39.65\%$   & $ 52.28 \%$   \\ \hline 	 	 
	\end{tabular}
\end{table}

\subsection{APPLICATIONS}
\textbf{Recovering structure from partial occupancy.}\\
In our dataset, we included partial occupancy at different levels and by the nature of our sampling we will  re-visit and propose occupancy changes with higher probability at the top levels. In the context of sketches, these examples correspond to partially observed structure - unfinished drawings - for which, given enough copies, the intended structure can be inferred and be employed to make suggestion or automatic fill-ins. Some examples of such samples can be viewed in Table \ref{tab:partialoccup}. We also report the sample with the highest likelihood, $I_{ML}$, and the sample with the highest occurrence in the posterior, $I_{MAP}$. This is not a simple inference problem: usually the full structure is inferred before a perfect recovery is achieved, thus the sampler has to be quite confident in the discovered structure in order to overcome not explaining fully the data, which is penalized by the likelihood. This trade-off is mediated by the noise instances and the blurring intensity. Once the model starts to match well parts of the input, the blur width and variance start to decrease, which increases the likelihood. This is only possible if the noise instances $\mathcal{N}$ match well the perturbations present in the input. This is why accounting for $\mathcal{N}$ in the model was found to be essential for hand-drawn samples.

\begin{table}[htc]
		\caption{Recovering structure from partial occupancy}
		\label{tab:partialoccup}
	\begin{tabular}{p{1.cm} |c c| c}
 \hline \\ [-0.3cm]
	\multicolumn{1}{ c| }{Input($I_D$)} & \multicolumn{2}{ |c }{Inferred image} &  \multicolumn{1}{ |c }{Inferred Structure}\\ \hline
	& $I_{MAP}$ & $I_{ML}$ & Possible expressions \\
	\fbox{\includegraphics[width=0.08\columnwidth]{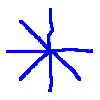}} &
			\fbox{\includegraphics[width=0.08\columnwidth]{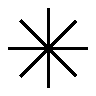}}  & 
		\fbox{\includegraphics[width=0.08\columnwidth]{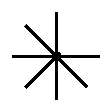}} & $ [(\ttranslate\ \ty),(\trotate\ 8)] $ \\ &
	\fbox{\includegraphics[width=0.08\columnwidth]{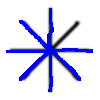}}  & 
		\fbox{\includegraphics[width=0.08\columnwidth]{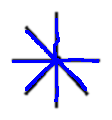}} & 
		$ [(\ttranslate\ \tx),(\trotate\ 8)] $ \\
		\hline
		\fbox{\includegraphics[width=0.08\columnwidth]{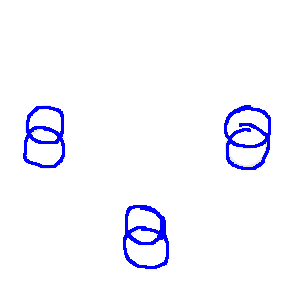}} &
			\fbox{\includegraphics[width=0.08\columnwidth]{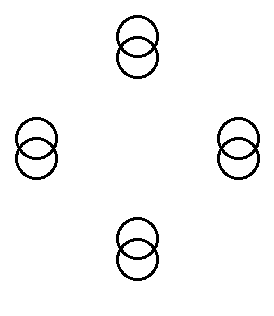}}  & 
		\fbox{\includegraphics[width=0.08\columnwidth]{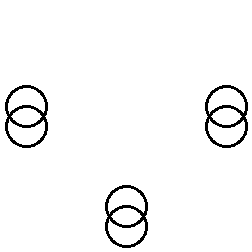}} & 
		\multirow{2}{*}{\parbox{3cm}{$ [(\ttranslate\ \tx),(\trotate\ 2\pi)$, 
		$(\ttranslate\ \tx), (\trotate\ 4),$ 
		$ (\ttranslate\ \ty)] $}} \\ &
	\fbox{\includegraphics[width=0.08\columnwidth]{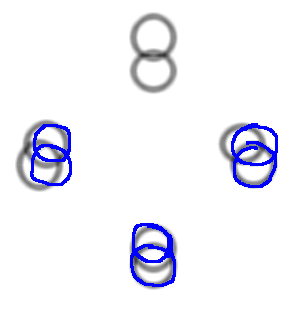}}  & 
		\fbox{\includegraphics[width=0.08\columnwidth]{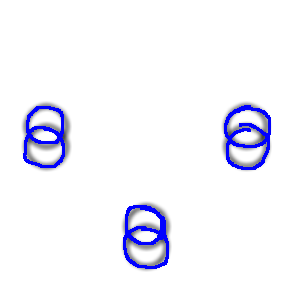}} & \\
		\hline
	\end{tabular}
\end{table}

\textbf{Common regular structures}\\
As most of the examples sampled from the prior looked rather abstract, expecially the ones with more complex structure, we look at some more common regular structures. First, we look into recovering regular polygons. We have already seen the example of the square and we can express any regular polygon in a similar fashion. One way of doing it is: starting with the origin, we translate it horizontally to make a line, then translate the line vertically, and then perform the $n$-fold rotation. Below is the description in our grammar:
$
     (\ttranslate\ \tx, [-t,t]),(\ttranslate\ \ty, [h]),
     (\trotate\ n, [0..n-1])
$.\\
The above control group, applied to the origin, will produce an $n$-sided regular polygon, of length $L = 2t$ and height $h$. The structure is quite simple and arises naturally in various samples from the prior, but exact recovery is a more challenging problem as we have a strong preference for occupancy that represents integer multiples of the scale unit $\lambda$. In general, we found that this is a reasonable assumption, but in this particular case there is a deterministic relation between $L$ and $h$ and in most of the case there is no choice of $\lambda$ that will assure both $L$ and $h$ to be integer multiples of such a unit. 
\begin{table}[h]
	\vskip -0.1in
	\caption{Different $h  = 1..3$, keeping the side length fixed}
	\begin{tabular}{c l}
	\fbox{\includegraphics[width=0.15\columnwidth]{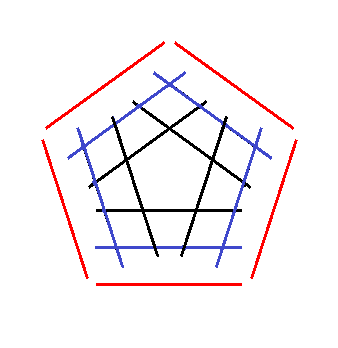}}
	& \parbox[3cm]{3cm}{$[ \begin{array}[t]{@{}l}
     (\ttranslate\ \tx, [-3,3]),(\ttranslate\ \ty, [-h]),\\(\trotate\ 5, [0..4])]
     \end{array} $} \\ 
	\end{tabular}
\end{table}
The recovery of the rotation and translation symmetries is not affected, but the deterministic/constraint relationship between $L$ and $h$ of exact regular polygons actually encodes additional structure. This can be captured by a slight change to the control group\footnote{We define the scaling group $G_S(\lambda) = \{\lambda^kI_2| k \in \mathcal{I} = \mathbb{Z} \}$}:
\begin{equation} \label{eq:regularPolygon}
	 \begin{array}[t]{@{}l}
	 (\tscale\ l, [-1]), (\ttranslate\ \tx, [-t,t]),
     (\tscale\ l, [1]), \\
     (\ttranslate\ \ty, [h]),
     (\trotate\ n, [0..n-1])]
  \end{array}
\end{equation}
  where $l = (2t)^{-1}h\tan{\pi/n}$. The control group above can be applied to any other fiber group to create a regular polygon using this fiber group as the building block. 

  
\begin{table}[htc]
		\caption{Regular polygons}
	\begin{tabular}{p{0.8cm}|cccc}
 	\hline \\ [-0.3cm]
 	
 	Input &
		\fbox{\includegraphics[width=0.1\columnwidth]{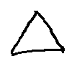}}  &
		\fbox{\includegraphics[width=0.1\columnwidth]{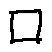}}  &
		\fbox{\includegraphics[width=0.1\columnwidth]{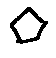}}  &
		\fbox{\includegraphics[width=0.1\columnwidth]{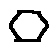}} \\
		
		\hline
	$I_{MAP}$ &
		\fbox{\includegraphics[width=0.1\columnwidth]{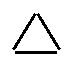}}  &
		\fbox{\includegraphics[width=0.1\columnwidth]{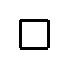}}  &
		\fbox{\includegraphics[width=0.1\columnwidth]{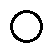}}  &
		\fbox{\includegraphics[width=0.1\columnwidth]{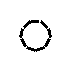}} \\
	
	$I_{ML}$ &
		\fbox{\includegraphics[width=0.1\columnwidth]{Figures/RegularPolygons/Drawn_0/exact_15_30000.png}}  &
		\fbox{\includegraphics[width=0.1\columnwidth]{Figures/RegularPolygons/Drawn_4/exact_0_30000.png}}  &
		\fbox{\includegraphics[width=0.1\columnwidth]{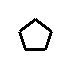}}  &
		\fbox{\includegraphics[width=0.1\columnwidth]{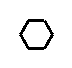}} \\

	\hline
	\end{tabular}
\end{table}

\par \textbf{Predefined Grid-like structures.} \\
 We also tried out some common regular structures (Table \ref{tab:grids}), like grids, having regular polygons as fiber groups. To speed up the inference, we predefined the regular polygon control group described in Eq. \ref{eq:regularPolygon} and used it in the proposal mechanism, as preferred structure. We report the shape scoring the maximum likelihood and the one with the highest posterior. Inference will prefer simpler explanations, as it can be seen from the third and fourth example. The complexity of the original shapes is slightly higher, but it explains well actual fiber copies in the input, and in fact most of the pixels in the input. In principle, we could force the likelihood to penalize more unexplained pixels, if full recoverability if important.  But as seen before a more forgiving likelihood allows for inference of structure with unobserved copies. Depending on the application, one would need to trade complexity and fidelity to the input, $I_D$.

\begin{table}[htc]
	\vskip -0.1in
		\caption{Predefined regular structures}
		\label{tab:grids}
	\begin{tabular}{p{0.8cm}|ccccc}
 	\hline \\ [-0.3cm]
 	
 	Input &
		\fbox{\includegraphics[width=0.09\columnwidth]{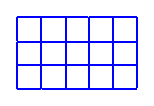}}  &
		\fbox{\includegraphics[width=0.09\columnwidth]{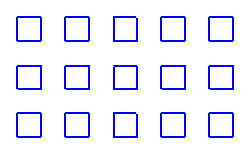}}  &
		\fbox{\includegraphics[width=0.09\columnwidth]{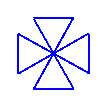}}  & 
		\fbox{\includegraphics[width=0.09\columnwidth]{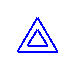}}&
		\fbox{\includegraphics[width=0.09\columnwidth]{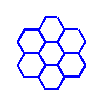}}\\
		
		\hline
	$I_{MAP}$ &
		\fbox{\includegraphics[width=0.1\columnwidth]{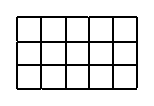}}  &
		\fbox{\includegraphics[width=0.1\columnwidth]{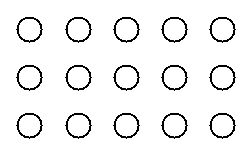}}  &
		\fbox{\includegraphics[width=0.09\columnwidth]{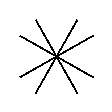}}& 
		\fbox{\includegraphics[width=0.09\columnwidth]{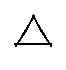}}&
		\fbox{\includegraphics[width=0.09\columnwidth]{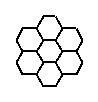}}\\
	
	$I_{ML}$ &
		\fbox{\includegraphics[width=0.09\columnwidth]{Figures/Grids/GT_3/exact_3_30000.png}}  &
		\fbox{\includegraphics[width=0.09\columnwidth]{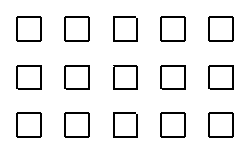}}  &
		\fbox{\includegraphics[width=0.09\columnwidth]{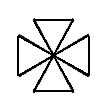}}& 
		\fbox{\includegraphics[width=0.09\columnwidth]{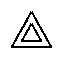}}&
		\fbox{\includegraphics[width=0.09\columnwidth]{Figures/Grids/GT_14/exact_1_30000.png}}\\

	\hline
	\end{tabular}
\end{table}

\textbf{Architectural sketches: Floor Plans}\\
Lastly, we download floor plans sketches of two famous buildings: the \textit{Dome of the Rock} \footnote{\url{http://en.wikipedia.org/wiki/Dome_of_the_Rock}}, representative of Islamic architecture and the \textit{Villa La Rotonda} \footnote{\url{http://en.wikipedia.org/wiki/Villa_Rotonda}}, landmark of Palladian architecture, as presented in Table \ref{tab:architecture}.

\begin{table}[h]
	\vskip -0.2in
	\caption{Floor plans inference results}
	 \label{tab:architecture}
	\begin{tabular}{c|ccccc}
	\hline
	\multicolumn{1}{c|}{Input($I_D$)} & \multicolumn{5}{c}{Samples from the posterior} \\ \hline
	
	\fbox{\includegraphics[width=0.1\columnwidth]{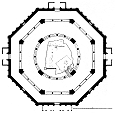}} &
	\fbox{\includegraphics[width=0.09\columnwidth]{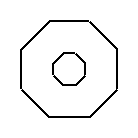}} &
	\fbox{\includegraphics[width=0.09\columnwidth]{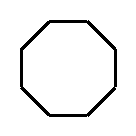}} &
	\fbox{\includegraphics[width=0.09\columnwidth]{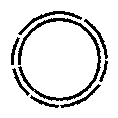}} &
	\fbox{\includegraphics[width=0.09\columnwidth]{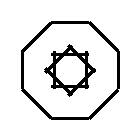}} &
	\fbox{\includegraphics[width=0.09\columnwidth]{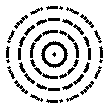}} \\
	
	&
	\fbox{\includegraphics[width=0.09\columnwidth]{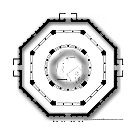}} &
	\fbox{\includegraphics[width=0.09\columnwidth]{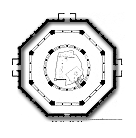}} &
	\fbox{\includegraphics[width=0.09\columnwidth]{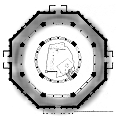}} &
	\fbox{\includegraphics[width=0.09\columnwidth]{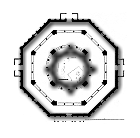}} &
	\fbox{\includegraphics[width=0.09\columnwidth]{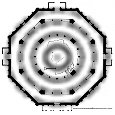}} \\

	\hline

	\fbox{\includegraphics[width=0.1\columnwidth]{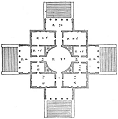}} &
	\fbox{\includegraphics[width=0.09\columnwidth]{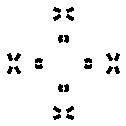}} &
	\fbox{\includegraphics[width=0.09\columnwidth]{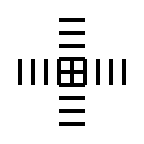}} &
	\fbox{\includegraphics[width=0.09\columnwidth]{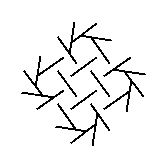}} &
	\fbox{\includegraphics[width=0.09\columnwidth]{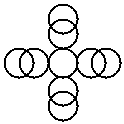}} &
	\fbox{\includegraphics[width=0.09\columnwidth]{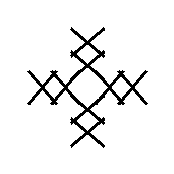}} \\
	&
	\fbox{\includegraphics[width=0.09\columnwidth]{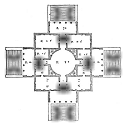}} &
	\fbox{\includegraphics[width=0.09\columnwidth]{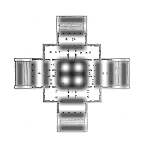}} &
	\fbox{\includegraphics[width=0.09\columnwidth]{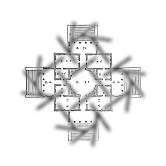}} &
	\fbox{\includegraphics[width=0.09\columnwidth]{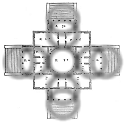}} &
	\fbox{\includegraphics[width=0.09\columnwidth]{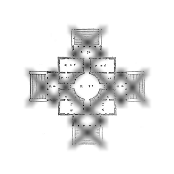}} \\
	
	\hline

%
	\end{tabular}
\end{table}

\section{RELATED WORK}
Wreath products have been applied previously to computer vision and image processing, in particular for multiresolution analysis generalizing approaches based on Haar/Fourier transform \cite{Foote99awreath} \cite{Foote2004a}.
\\Treating vision as an inverse inference problem aims to estimate the causes and factors that describe a generative history - generally proposing some hierarchical representations. These usually employ a bottom-up generative process, coupled with some kind of top-down validation and have been successfully used in image and scene parsing, but they usually require expert knowledge in setting up the hierarchy and encoding a known high level structure(like spacial relationships between objects/primitives) \cite{Tu05imageparsing} \cite{Bottom-Up09}. In contrast, the wreath process automatically detects this structure by maximization of transfer. \\Most literature on sketch beautification often employs beautification by recognition: they provide a vocabulary of primitives and any object in the data must be represented in this vocabulary. The approach has limited generalization by itself. More recently, the idea of constructing more complex objects out of a group of easily detectable primitives was used in \cite{Paulson2008} \cite{Kara2004}. Note that such methods could be used in conjunction with the wreath process.

\pdfoutput=1
\section{CONCLUSION}
In summary, the three main contributions of this paper are:  We propose the stochastic wreath process as a new, highly structured random point process, thus generalizing Leyton's generative theory of shape; We propose an inference scheme for inferring structure and parameters of the wreath process for a given observed pixel image; We report on experimental results of the inference based on both model-generated as well as hand-drawn images of geometric shapes.

While our experiments were restricted to the domain of two dimensional monochromatic geometric figures, the same kind of hierarchical generative model can also be applied to three-dimensional shapes. Also, as mentioned in Leyton's book, the action of the group itself can be different from inking and could also include cutting away of material or similar shape-creating actions.


Finally, a wreath product representation can be viewed as providing a natural coordinate system for a shape in the most general sense. For example, in the case of the square, the wreath product representation provides a set of natural coordinates for every point on the square specifying which side the point is on, and where on that side it is located. In this sense, discovering the underlying wreath process of a shape can be understood as finding a meaningful coordinate system for describing parts of that shape. This principle can be generalized to other structures, including finite state automata, and the stochastic wreath process and associated inference might find applications in such other domains, for example in the analysis of genetic regulatory networks as outlined in \cite{Egri-Nagy2008}.

%

\bibliography{references}

\begin{thebibliography}{10}

\bibitem{Egri-Nagy2008}
Attila Egri-Nagy and Chrystopher~L Nehaniv.
\newblock Hierarchical coordinate systems for understanding complexity and its
  evolution, with applications to genetic regulatory networks.
\newblock {\em Artificial Life}, 14(3):299--312, 2008.

\bibitem{Foote2004a}
Richard Foote, Gagan Mirchandani, and Daniel Rockmore.
\newblock Two-dimensional wreath product group-based image processing.
\newblock {\em Journal of Symbolic Computation}, 37(2):187--207, 2004.

\bibitem{Foote99awreath}
Richard Foote, Gagan Mirchandani, Daniel~N Rockmore, Dennis Healy, and Tim
  Olson.
\newblock A wreath product group approach to signal and image processing. i.
  multiresolution analysis.
\newblock {\em Signal Processing, IEEE Transactions on}, 48(1):102--132, 2000.

\bibitem{Green95reversiblejump}
Peter~J. Green.
\newblock Reversible jump {Markov} chain {Monte Carlo} computation and
  {Bayesian} model determination.
\newblock {\em Biometrika}, 82:711--732, 1995.

\bibitem{Green09reversiblejump}
Peter~J Green and David~I Hastie.
\newblock Reversible jump mcmc.
\newblock {\em Genetics}, 155(3):1391--1403, 2009.

\bibitem{Bottom-Up09}
Feng Han and Song-Chun Zhu.
\newblock Bottom-up/top-down image parsing with attribute grammar.
\newblock {\em Pattern Analysis and Machine Intelligence, IEEE Transactions
  on}, 31(1):59--73, 2009.

\bibitem{Kara2004}
Levent~Burak Kara and Thomas~F. Stahovich.
\newblock Hierarchical parsing and recognition of hand-sketched diagrams.
\newblock In {\em Proceedings of the 17th annual ACM symposium on User
  interface software and technology}, UIST '04, pages 13--22, New York, NY,
  USA, 2004. ACM.

\bibitem{Leyton01Generative}
Michael Leyton.
\newblock {\em A Generative Theory of Shape}.
\newblock Number LNCS 2145 in Lecture Notes in Computer Science.
  Springer-Verlag, 2001.

\bibitem{Mansinghka2013}
Vikash Mansinghka, Tejas~D Kulkarni, Yura~N Perov, and Josh Tenenbaum.
\newblock Approximate bayesian image interpretation using generative
  probabilistic graphics programs.
\newblock In {\em Advances in Neural Information Processing Systems}, pages
  1520--1528, 2013.

\bibitem{Marjoram03}
Paul Marjoram, John Molitor, Vincent Plagnol, and Simon Tavaré.
\newblock Markov chain {Monte Carlo} without likelihoods.
\newblock {\em Proceedings of the National Academy of Sciences},
  100(26):15324--15328, 2003.

\bibitem{Paulson2008}
Brandon Paulson and Tracy Hammond.
\newblock Paleosketch: accurate primitive sketch recognition and
  beautification.
\newblock In {\em Proceedings of the 13th international conference on
  Intelligent user interfaces}, IUI '08, pages 1--10, New York, NY, USA, 2008.
  ACM.

\bibitem{Sisson07}
S.~A. Sisson, Y.~Fan, and Mark~M. Tanaka.
\newblock Sequential {Monte Carlo} without likelihoods.
\newblock {\em Proceedings of the National Academy of Sciences},
  104(6):1760--1765, 2007.

\bibitem{Tu05imageparsing}
Zhuowen Tu, Xiangrong Chen, Alan~L Yuille, and Song-Chun Zhu.
\newblock Image parsing: Unifying segmentation, detection, and recognition.
\newblock {\em International Journal of Computer Vision}, 63(2):113--140, 2005.

\bibitem{Wilkinson08}
R.~D. {Wilkinson}.
\newblock {Approximate {Bayesian} computation (ABC) gives exact results under
  the assumption of model error}.
\newblock {\em ArXiv e-prints}, November 2008.

\end{thebibliography}
\bibliographystyle{plain}

\end{document}